\def\cvprPaperID{0000} %
\def\confName{CVPR}
\def\confYear{2024}

\def\paperTitle{4K4D: Real-Time 4D View Synthesis at 4K Resolution}

\def\authorBlock{
    Zhen Xu\textsuperscript{1} \quad
    Sida Peng\textsuperscript{1} \quad
    Haotong Lin\textsuperscript{1} \quad
    Guangzhao He\textsuperscript{1} \quad \\[3pt]
    Jiaming Sun\textsuperscript{2} \quad
    Yujun Shen\textsuperscript{3} \quad
    Hujun Bao\textsuperscript{1} \quad
    Xiaowei Zhou\textsuperscript{1} \\[8pt]
    \textsuperscript{1}Zhejiang University \qquad
    \textsuperscript{2}Image Derivative Inc. \qquad
    \textsuperscript{3}Ant Group
}

\newif\ifreview 
\newif\ifarxiv \newcommand{\arxiv}{\arxivtrue}
\newif\ifcamera 
\newif\ifrebuttal 

\newcommand{\method}{\textcolor{black}{4K4D}\xspace}

\arxiv

\pdfoutput=1
\documentclass[10pt,twocolumn,letterpaper]{article}
\ifreview \usepackage[review]{cvpr} \fi
\ifarxiv \usepackage[pagenumbers]{cvpr} \fi
\ifrebuttal \usepackage[rebuttal]{cvpr} \fi
\ifcamera \usepackage{cvpr} \fi

\usepackage{graphicx}
\usepackage{amsmath}
\usepackage{amssymb}
\usepackage{booktabs}

\usepackage{times}
\usepackage{microtype}
\usepackage{epsfig}
\usepackage[table,xcdraw]{xcolor}
\usepackage{caption}
\usepackage{float}
\usepackage{placeins}
\usepackage{color, colortbl}
\usepackage{stfloats}
\usepackage{enumitem}
\usepackage{tabularx}
\usepackage{xstring}
\usepackage{multirow}
\usepackage{xspace}
\usepackage{url}
\usepackage{xcolor}
\usepackage{algorithm2e}
\usepackage[hang,flushmargin]{footmisc}

\ifcamera \usepackage[accsupp]{axessibility} \fi

\ifarxiv  \fi

\newcommand{\R}[1]{{%
    \textbf{%
        \ifstrequal{#1}{1}{\textcolor{red}{R#1}}{%
        \ifstrequal{#1}{2}{\textcolor{blue}{R#1}}{%
        \ifstrequal{#1}{3}{\textcolor{magenta}{R#1}}{%
        \ifstrequal{#1}{4}{\textcolor{teal}{R#1}}{%
                           \textcolor{cyan}{R#1}%
        }}}}%
    }%
}}

\usepackage{xcolor}
\definecolor{myred}{rgb}{0.8,0,0}
\definecolor{mygreen}{rgb}{0,0.8,0}

\usepackage{silence}
\usepackage{xr-hyper}

\makeatletter
\newcommand*{\addFileDependency}[1]{
  \typeout{(#1)}
  \@addtofilelist{#1}
  \IfFileExists{#1}{}{\typeout{No file #1.}}
}

\makeatother

\definecolor{citeblue}{RGB}{48,111,186}
\usepackage[pagebackref=false,breaklinks,colorlinks,citecolor=citeblue]{hyperref}
\usepackage[capitalize]{cleveref}
\crefname{section}{Sec.}{Secs.}
\crefname{table}{Tab.}{Tabs.}
\crefname{figure}{Fig.}{Figs.}
\definecolor{colorfirst}{rgb}{.866,.945, 0.831} %
\definecolor{colorsecond}{rgb}{1, 0.98, 0.83} %
\definecolor{colorthird}{rgb}{0.76, 0.87, 0.92} %

\newcommand{\cellfirst}{\cellcolor{colorfirst}}
\newcommand{\cellsecond}{\cellcolor{colorsecond}}

\begin{document}
\title{\paperTitle}
\author{\authorBlock}

\twocolumn[
    \maketitle
    \vspace{-2em}
    \begin{center}
    \captionsetup{type=figure}
    \vspace{4pt}
    \includegraphics[width=0.985\textwidth]{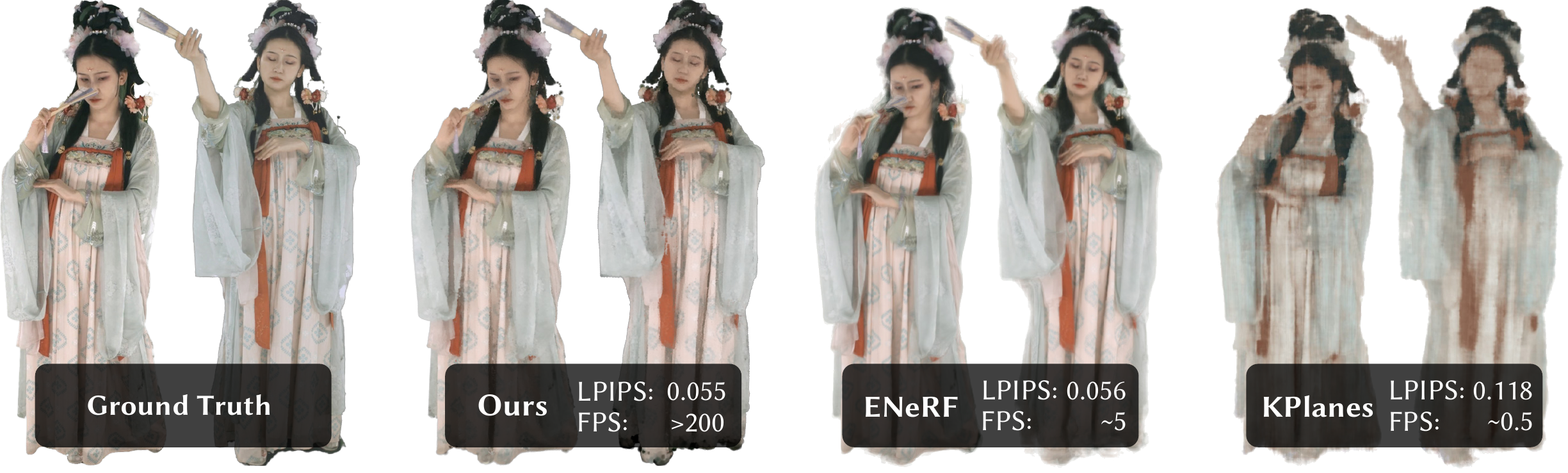}
    \vspace{0pt}
    \captionof{figure}{%
        \textbf{Photorealistic and real-time rendering of dynamic 3D scenes.}
        Our proposed method reconstructs a 4D neural representation from multi-view videos, which can be rendered at 1125$\times$1536 resolution with a speed of over 200~FPS using an RTX 3090 GPU while maintaining state-of-the-art quality on the DNA-Rendering~\cite{cheng2023dna} dataset.
        It is also noteworthy that our method reaches over 80~FPS when rendering 4K images with an RTX 4090.
        Detailed performance under different resolutions using different GPUs can be found in \cref{tab:ablation_resolution}.
    }
    \vspace{6pt}
    \label{fig:teaser}
\end{center}
    \bigbreak
]

\begin{abstract}
This paper targets high-fidelity and real-time view synthesis of dynamic 3D scenes at 4K resolution.
Recently, some methods on dynamic view synthesis have shown impressive rendering quality.
However, their speed is still limited when rendering high-resolution images.
To overcome this problem, we propose \method, a 4D point cloud representation that supports hardware rasterization and enables unprecedented rendering speed.
Our representation is built on a 4D feature grid so that the points are naturally regularized and can be robustly optimized.
In addition, we design a novel hybrid appearance model that significantly boosts the rendering quality while preserving efficiency.
Moreover, we develop a differentiable depth peeling algorithm to effectively learn the proposed model from RGB videos.
Experiments show that our representation can be rendered at over 400 FPS on the DNA-Rendering dataset at 1080p resolution and 80 FPS on the ENeRF-Outdoor dataset at 4K resolution using an RTX 4090 GPU, which is 30$\times$ faster than previous methods and achieves the state-of-the-art rendering quality.
Our project page is available at \href{https://zju3dv.github.io/4k4d/}{https://zju3dv.github.io/4k4d}.
\end{abstract}
\vspace{-10pt}

\section{Introduction}

Dynamic view synthesis aims to reconstruct dynamic 3D scenes from captured videos and create immersive virtual playback, which is a long-standing research problem in computer vision and computer graphics.
Essential to the practicality of this technique is its ability to be rendered in real time with high fidelity, enabling its application in VR/AR, sports broadcasting, and artistic performance capturing.
Traditional methods~\cite{hilsmann2020going,casas20144d,collet2015high,dou2016fusion4d,newcombe2015dynamicfusion,orts2016holoportation,yu2018doublefusion} represent dynamic 3D scenes as textured mesh sequences and reconstruct them using complicated hardware.
Therefore, they are typically limited to controlled environments.

Recently, implicit neural representations~\cite{mildenhall2021nerf,fridovich2023k,li2022neural} have shown great success in reconstructing dynamic 3D scenes from RGB videos via differentiable rendering.
For example, Li~\etal~\cite{li2022neural} model the target scene as a dynamic radiance field and leverage volume rendering~\cite{drebin1988volume} to synthesize images, which are compared with input images for optimization.
Despite impressive dynamic view synthesis results, existing approaches typically require seconds or even minutes to render an image at 1080p resolution due to the costly network evaluation, as discussed by Peng~\etal~\cite{peng2023representing}.

Inspired by static view synthesis approaches~\cite{yu2021plenoctrees,garbin2021fastnerf,kerbl20233d}, some dynamic view synthesis methods~\cite{peng2023representing,wang2022fourier,attal2023hyperreel} increase the rendering speed by decreasing either the cost or the number of network evaluations.
With these strategies, MLP Maps~\cite{peng2023representing} is able to render foreground dynamic humans with a speed of 41.7 fps.
However, the challenge of rendering speed still exists, since the real-time performance of MLP Maps is achieved only when synthesizing moderate-resolution images (384$\times$512).
When rendering 4K resolution images, its speed reduces to only 1.3 FPS.

In this paper, we propose a novel neural representation, named \method, for modeling and rendering dynamic 3D scenes.
As illustrated in \cref{fig:teaser}, \method significantly outperforms previous dynamic view synthesis approaches~\cite{lin2022efficient, fridovich2023k} in terms of the rendering speed, while being competitive in the rendering quality.
Our core innovation lies in a 4D point cloud representation and a hybrid appearance model.
Specifically, for the dynamic scene, we obtain the coarse point cloud sequence using a space carving algorithm~\cite{kutulakos2000theory} and model the position of each point as a learnable vector.
A 4D feature grid is introduced for assigning a feature vector to each point, which is fed into MLP networks to predict the point’s radius, density, and spherical harmonics (SH) coefficients~\cite{muller2006spherical}.
The 4D feature grid naturally applies spatial regularization on the point clouds and makes the optimization more robust, as supported by the results in \cref{sec:ablation}.
Based on \method, we develop a differentiable depth peeling algorithm that exploits the hardware rasterizer to achieve unprecedented rendering speed.

We find that the MLP-based SH model struggles to represent the appearance of dynamic scenes.
To alleviate this issue, we additionally introduce an image blending model to incorporate with the SH model to represent the scene's appearance.
An important design is that we make the image blending network independent from the viewing direction, so it can be pre-computed after training to boost the rendering speed.
As a two-edged sword, this strategy makes the image-blending model discrete along the viewing direction.
This problem is compensated for using the continuous SH model.
In contrast to 3D Gaussian Splatting~\cite{kerbl20233d} that uses the SH model only, our hybrid appearance model fully exploits the information captured by input images, thus effectively improving the rendering quality.

To validate the effectiveness of the proposed pipeline, we evaluate \method on multiple widely used datasets for multi-view dynamic novel view synthesis, including NHR~\cite{wu2020multi}, ENeRF-Outdoor~\cite{lin2022efficient}, DNA-Rendering~\cite{cheng2023dna}, and Neural3DV~\cite{li2021neural}.
Extensive experiments show that \method could not only be rendered orders of magnitude faster but also notably outperform the state-of-the-art in terms of rendering quality.
With an RTX 4090 GPU, our method reaches 400 FPS on the DNA-Rendering dataset at 1080p resolution and 80 FPS on the ENeRF-Outdoor dataset at 4K resolution.

\begin{figure*}[t]
    \centering
    \includegraphics[width=0.985\textwidth]{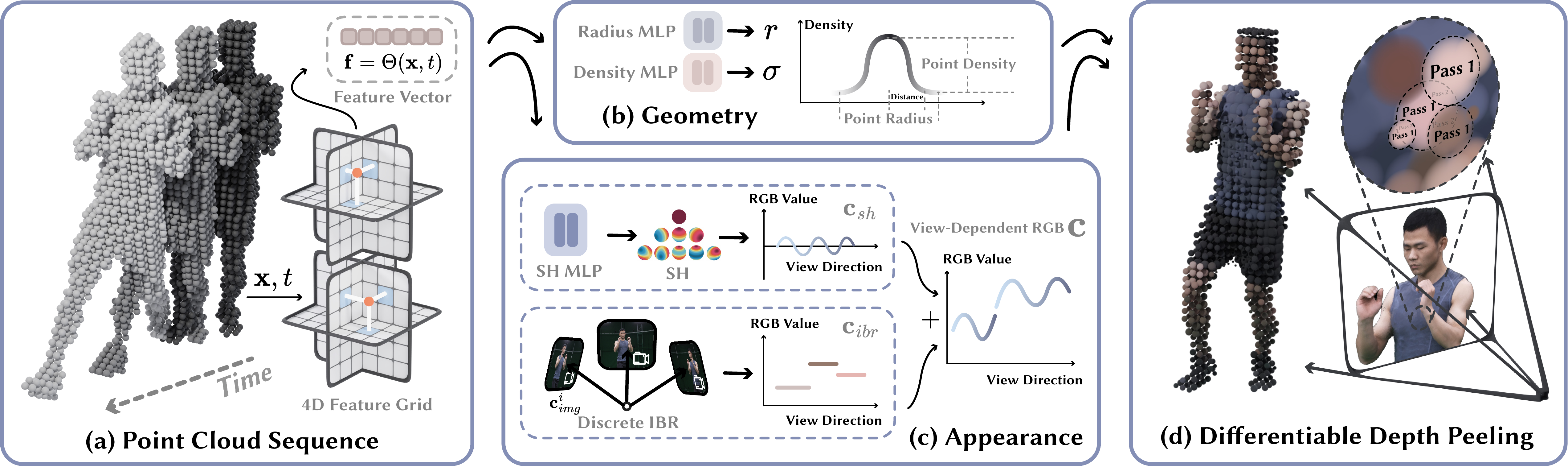}
    \caption{
        \textbf{Overview of our proposed pipeline.} %
        (a) By applying the space-carving algorithm \cite{kutulakos2000theory}, we extract the initial cloud sequence $\mathbf{x}, t$ of the target scene.
        A 4D feature grid \cite{fridovich2023k} is predefined to assign a feature vector to each point, which is then fed into MLPs for the scene geometry and appearance.
        (b) The geometry model is based on the point location, radius, and density, which forms a semi-transparent point cloud.
        (c) The appearance model consists of a piece-wise constant IBR term $\mathbf{c}_{ibr}$ and a continuous SH model $\mathbf{c}_{sh}$.
        (d) The proposed representation is learned from multi-view RGB videos through the differentiable depth peeling algorithm.
    }
    \label{fig:pipeline}
\end{figure*}

\section{Related Work}

\noindent\textbf{Neural scene representations.}
In the domain of novel view synthesis, various approaches have been proposed to address this challenging problem,
including multi-view image-based methods \cite{chris2001unstructured,chaurasia2013depth,flynn2016deepstereo,kalantari2016learning,penner2017soft,zitnick2004high},
multi-plane image representations \cite{li2020crowdsampling,mildenhall2019local,srinivasan2019pushing,szeliski1998stereo,parker1998single,szeliski1998stereo},
light-field techniques \cite{davis2012unstructured,gortler1996lumigraph,levoy1996light}
as well as explicit surface or voxel-based methods \cite{collet2015high,dou2016fusion4d,newcombe2015dynamicfusion,orts2016holoportation,yu2018doublefusion}.
\cite{collet2015high} utilizes depth sensors and multi-view stereo techniques to consolidate per-view depth information into a coherent scene geometry, producing high-quality volumetric video.
These methods require intricate hardware setups and studio arrangements, thus constraining their accessibility and applicability.
Recently, implicit neural scene representations\cite{attal2022learning,hedman2018deep,jiang2020sdfdiff,kellnhofer2021neural,liu2019learning,lombardi2019neural,shih20203d,sitzmann2021light,sitzmann2019deepvoxels,sitzmann2019scene,suhail2022cvpr,wizadwongsa2021nex,mildenhall2021nerf} have attracted significant interest among researchers.
NeRF\cite{mildenhall2021nerf} encodes the radiance fields of static scenes using coordinate-based Multi-Layer Perceptrons (MLP), achieving exceptional novel view synthesis quality.

Building upon NeRF, a collection of studies \cite{li2022neural,pumarola2020d,park2021hypernerf,park2021nerfies,li2020neural,wu2020multi,icsik2023humanrf} has extended implicit neural representations to accommodate dynamic scenes.
DyNeRF \cite{li2022neural} introduces an additional temporal dimension to NeRF's 5D input, thereby enabling it to model temporal variations in dynamic scenes.
However, NeRF-based approaches often suffer from substantial computational costs, leading to rendering times of seconds or even minutes for moderate-resolution images, which significantly limits their practicality.
Another line of studies \cite{wang2021ibrnet,yu2020pixelnerf,chen2021mvsnerf,li2023dynibar} has concentrated on integrating image features into the NeRF rendering pipeline.
This approach is easily applicable to dynamic scenes, as multi-view videos can be effortlessly decomposed into multi-view images.
Nevertheless, the convolution operations employed in these methods, also face challenges in terms of rendering speed as the input image resolution increases, hindering the rendering efficiency of these approaches in real-world applications.

\vspace{2pt}
\noindent\textbf{Accelerating neural scene representations.}
Multiple studies have focused on accelerating the rendering speed of implicit neural scene representation by distilling implicit MLP networks into explicit structures that offer fast query capabilities,
including voxel grids \cite{yu2021plenoctrees,garbin2021fastnerf,hedman2021baking,yu2021plenoxels,reiser2021kilonerf,muller2022instant,li2023nerfacc},
explicit surfaces \cite{chen2022mobilenerf,kulhanek2023tetra,hasselgren2022shape,patakin2023neural,lu2023urban,jena2022neural}
and point-based representations \cite{kerbl20233d,kopanas2021point,zhang2022differentiable,ruckert2022adop,aliev2020neural,rakhimov2022npbg++,lassner2021pulsar}.
These methods effectively reduce the cost or the number of NeRF’s MLP evaluations required.
One notable advancement is the development of 3D Gaussian Splatting (3DGS) \cite{kerbl20233d} which introduces a differentiable splatting algorithm for differentiable volume rendering \cite{bavoil2008order,drebin1988volume}.
This technique leverages semi-transparent Gaussian ellipsoids with spherical harmonics \cite{muller2006spherical} to attain both high-fidelity and high-speed rendering,
effectively eliminating the slow ray marching operation.
However, the aforementioned acceleration techniques are only applicable to static scenes.

Inspired by the success of techniques for accelerating static neural representations,
several approaches \cite{attal2023hyperreel,lombardi2021mixture,wang2022mixed,lin2022efficient,wang2023neural,song2023nerfplayer,peng2023representing,lin2023im4d,chen2022tensorf,shao2022tensor4d,song2023nerfplayer} have explored the possibility of real-time dynamic view synthesis.
HyperReel \cite{attal2023hyperreel} employs a primitive prediction module to reduce the number of network evaluations,
thereby achieving real-time rendering speeds at moderate resolutions.
However, it should be noted that their rendering speed decreases significantly when rendering higher-resolution images,
as evidenced by experiments detailed in \cref{sec:experiments} (\textit{e.g.}, 1.5FPS for a 1352$\times$1014 image from the Neural3DV \cite{li2022neural} dataset).
In recent developments, a line of concurrent work \cite{luiten2023dynamic,wu20234d,yang2023realtime,yang2023deformable} has also reported real-time rendering speeds by incorporating correspondence or time-dependency into the 3DGS approach \cite{kerbl20233d}.
However, these methods do not show results on datasets with large and fast motions (DNA-Rendering \cite{cheng2023dna} and NHR \cite{wu2020multi}) and could only achieve real-time speed at moderate resolution (800$\times$800 \cite{wu20234d} and 640$\times$480 \cite{luiten2023dynamic}).
In contrast, \method is capable of achieving real-time rendering even at 4K resolution while concurrently maintaining state-of-the-art view-synthesis quality on large-motion data (as elaborated in \cref{sec:experiments}).

\section{Proposed Approach}

Given a multi-view video capturing a dynamic 3D scene, our goal is to reconstruct the target scene and perform novel view synthesis in real time.
To this end, we extract coarse point clouds of the scene using the space-carving algorithm \cite{kutulakos2000theory} (\cref{sec:implementation}) and build a point cloud-based neural scene representation, which can be robustly learned from input videos and enable the hardware-accelerated rendering.

The overview of the proposed model is presented in \cref{fig:pipeline}.
In this section, we first describe how to represent the geometry and appearance of dynamic scenes based on point clouds and neural networks (\cref{sec:representation}).
Then, we develop a differentiable depth peeling algorithm for rendering our representation (\cref{sec:rendering}), which is supported by the hardware rasterizer, thereby significantly improving the rendering speed.
Finally, we discuss how to optimize the proposed model on input RGB videos (\cref{sec:training}).

\subsection{Modeling Dynamic Scenes with Point Clouds}\label{sec:representation}

\noindent\textbf{4D embedding.}  %
Given the coarse point clouds of the target scene, we represent its dynamic geometry and appearance using neural networks and feature grids.
Specifically, our method first defines six feature planes $\theta_{xy}$, $\theta_{xz}$, $\theta_{yz}$, $\theta_{tx}$, $\theta_{ty}$, and $\theta_{tz}$.
To assign a feature vector $\mathbf{f}$ to any point $\mathbf{x}$ at frame $t$, we adopt the strategy of K-Planes \cite{fridovich2023k} to model a 4D feature field $\Theta(\mathbf{x}, t)$ using these six planes:
\begin{multline}
    \label{eq:sample_feture_volume}
    \mathbf{f} = \Theta(\mathbf{x}, t) =
    \theta_{xy}(x, y)\oplus\theta_{xz}(x, z)\oplus\theta_{yz}(y, z)\oplus \\
    \theta_{tx}(t, x)\oplus\theta_{ty}(t, y)\oplus\theta_{tz}(t, z),
\end{multline}
where $\mathbf{x} = (x, y, z)$ is the input point, and $\oplus$ indicates the concatenation operator.
Please refer to K-Planes \cite{fridovich2023k} for more implementation details.

\vspace{2pt}
\noindent\textbf{Geometry model.}  %
Based on coarse point clouds, the dynamic scene geometry is represented by learning three entries on each point: position $\mathbf{p} \in R^3$, radius $r \in R$, and density $\sigma \in R$.
Using these point entries, we calculate the volume density of space point $\mathbf{x}$ with respect to an image pixel $\mathbf{u}$ for the volume rendering, which will be described in \cref{sec:rendering}.
The point position $\mathbf{p}$ is modeled as an optimizable vector.
The radius $r$ and density $\sigma$ are predicted by feeding the feature vector $\mathbf{f}$ in \cref{eq:sample_feture_volume} to an MLP network.

\vspace{2pt}
\noindent\textbf{Appearance model.}  %
As illustrated in \cref{fig:pipeline}c, we use the image blending technique and the spherical harmonics (SH) model \cite{muller2006spherical, yu2021plenoctrees} to build a hybrid appearance model, where the image blending technique represents the discrete view-dependent appearance $\mathbf{c}_{ibr}$ and the SH model represents the continuous view-dependent appearance $\mathbf{c}_{sh}$.
For point $\mathbf{x}$ at frame $t$, its color with viewing direction $\mathbf{d}$ is:
\begin{equation}
    \label{eq:appearance}
    \mathbf{c}(\mathbf{x}, t, \mathbf{d}) = \mathbf{c}_{ibr}(\mathbf{x}, t, \mathbf{d}) + \mathbf{c}_{sh}(\mathbf{s}, \mathbf{d}),
\end{equation}
where $\mathbf{s}$ means SH coefficients at point $\mathbf{x}$.

The discrete view-dependent appearance $\mathbf{c}_{ibr}$ is inferred based on input images.
Specifically, for a point $\mathbf{x}$, we first project it into the input image to retrieve the corresponding RGB color $\mathbf{c}_{img}^i$.
Then, to blend input RGB colors, we calculate the corresponding blending weight $w^i$ based on the point coordinate and the input image.
Note that the blending weight is independent from the viewing direction.
Next, to achieve the view-dependent effect, we select the $N'$ nearest input views according to the viewing direction.
Finally, the color $\mathbf{c}_{ibr}$ is computed as $\sum_{i=1}^{N'} w^i \mathbf{c}_{img}^i$.
Because the $N'$ input views are obtained through the nearest neighbor retrieval, the $\mathbf{c}_{ibr}$ is inevitably discrete along the viewing direction.
To achieve the continuous view-dependent effect, we append the fine-level color $\mathbf{c}_{sh}$ represented by the SH model, as shown in \cref{fig:pipeline}c.

In practice, our method regresses the SH coefficients $\mathbf{s}$ by passing the point feature $\mathbf{f}$ in \cref{eq:sample_feture_volume} into an MLP network.
To predict the blending weight $w^i$ in the image blending model $\mathbf{c}_{ibr}$, we first project point $\mathbf{x}$ onto the input image to retrieve the image feature $\mathbf{f}_{img}^i$, and then concatenate it with the point feature $\mathbf{f}$, which is fed into another MLP network to predict the blending weight.
The image feature $\mathbf{f}_{img}^i$ is extracted using a 2D CNN network.

\vspace{2pt}
\noindent\textbf{Discussion.}  %
Our appearance model is the key to achieving the low-storage, high-fidelity, and real-time view synthesis of dynamic scenes.
There are three alternative ways to represent the dynamic appearance, but they cannot perform on par with our model.
1) Defining explicit SH coefficients on each point, as in 3D Gaussian splatting \cite{kerbl20233d}.
When the dimensional of SH coefficients is high and the amount of points of dynamic scenes is large, this model's size could be too big to train on a consumer GPU.
2) MLP-based SH model.
Using an MLP to predict SH coefficients of each point can effectively decrease the model size.
However, our experiments found that MLP-based SH model struggles to render high-quality images (\cref{sec:ablation}).
3) Continuous view-dependent image blending model, as in ENeRF \cite{lin2022efficient}.
We found that representing the appearance with the image blending model has better rendering quality than with only MLP-based SH model.
However, the network in ENeRF takes the viewing direction as input and thus cannot be easily pre-computed, limiting the rendering speed during inference.

In contrast to these three methods, our appearance model combines a discrete image blending model $\mathbf{c}_{ibr}$ with a continuous SH model $\mathbf{c}_{sh}$.
The image blending model $\mathbf{c}_{ibr}$ boosts the rendering performance.
In addition, it supports the pre-computation, as its network does not take the viewing direction as input.
The SH model $\mathbf{c}_{sh}$ enables the view-dependent effect for any viewing direction.
During training, our model represents the scene appearance using networks, so its model size is reasonable.
During inference, we pre-compute the network outputs to achieve the real-time rendering, which will be described in \cref{sec:inference}.

\subsection{Differentiable Depth Peeling}
\label{sec:rendering}

Our proposed dynamic scene representation can be rendered into images using the depth peeling algorithm \cite{bavoil2008order}.
Thanks to the point cloud representation, we are able to leverage the hardware rasterizer to significantly speed up the depth peeling process.
Moreover, it is easy to make this rendering process differentiable, enabling us to learn our model from input RGB videos.

We develop a custom shader to implement the depth peeling algorithm that consists of $K$ rendering passes.
Consider a particular image pixel $\mathbf{u}$.
In the first pass, our method first uses the hardware rasterizer to render point clouds onto the image, which assigns the closest-to-camera point $\mathbf{x}_0$ to the pixel $\mathbf{u}$.
Denote the depth of point $\mathbf{x}_0$ as $t_0$.
Subsequently, in the $k$-th rendering pass, all points with depth value $t_k$ smaller than the recorded depth of the previous pass $t_{k-1}$ are discarded, thereby resulting in the $k$-th closest-to-camera point $\mathbf{x}_k$ for the pixel $\mathbf{u}$.
Discarding closer points is implemented in our custom shader, so it still supports the hardware rasterization.
After $K$ rendering passes, pixel $\mathbf{u}$ has a set of sorted points $\{\mathbf{x}_k | k=1, ..., K\}$.

Based on the points $\{\mathbf{x}_k | k=1, ..., K\}$, we use the volume rendering to synthesize the color of pixel $\mathbf{u}$.
The densities of points $\{\mathbf{x}_k | k=1, ..., K\}$ for pixel $\mathbf{u}$ is defined based on the distance between the projected point and pixel $\mathbf{u}$ on the 2D image:
\begin{equation}
    \label{eq:density}
    \alpha(\mathbf{u}, \mathbf{x}) = \sigma \cdot \mathrm{max}(1 - \frac{||\pi(\mathbf{x}) - \mathbf{u}||_2^2}{r^2}, 0),
\end{equation}
where $\pi$ is the camera projection function. $\sigma$ and $r$ are the density and radius of point $\mathbf{x}$, which are described in \cref{sec:representation}.
During training, we implement the projection function $\pi$ using the PyTorch \cite{paszke2019pytorch}, so \cref{eq:density} is naturally differentiable.
During inference, we leverage the hardware rasterization process to efficiently obtain the distance $||\pi(\mathbf{x}) - \mathbf{u}||_2^2$, which is implemented using OpenGL \cite{shreiner2009opengl}.

Denote the density of point $\mathbf{x}_k$ as $\alpha_k$. The color of pixel $\mathbf{u}$ from the volume rendering is formulated as:
\begin{equation}
    \label{eq:volume_rendering}
    C(\mathbf{u}) = \sum_{k=1}^{K} T_k \alpha_k \mathbf{c}_k \text{, where } T_k = \prod_{j=1}^{k-1} (1 - \alpha_j),
\end{equation}
where $\mathbf{c}_k$ is the color of point $\mathbf{x}_k$, as described in \cref{eq:appearance}.

\subsection{Training}
\label{sec:training}
Given the rendered pixel color $C(\mathbf{u})$, we compare it with the ground-truth pixel color $C_{gt}(\mathbf{u})$ to optimize our model in an end-to-end fashion using the following loss function:
\begin{equation}
    \label{eq:rgb_loss}
    L_{img} = \sum_{\mathbf{u} \in \mathcal{U}} ||C(\mathbf{u}) - C_{gt}(\mathbf{u})||_2^2,
\end{equation}
where $\mathcal{U}$ is the set of image pixels.
In addition to the MSE loss $L_{img}$, we also apply the perceptual loss $L_{lpips}$ \cite{zhang2018unreasonable}.
\begin{equation}
    L_{lpips} = ||\Phi(I) - \Phi(I_{gt})||_1,
\end{equation}
where $\Phi$ is the perceptual function (a VGG16 network) and $I,I_{gt}$ are the rendered and ground-truth images, respectively.
The perceptual loss \cite{zhang2018unreasonable} computes the difference in image features extracted from the VGG model \cite{simonyan2014very}.
Our experiments in \cref{sec:ablation} show that it effectively improves the perceived quality of the rendered image.

To regularize the optimization process of our proposed representation, we additionally apply the mask supervision to dynamic regions of the target scene.
We solely render point clouds of dynamic regions to obtain their masks, where the pixel value is obtained by:
\begin{equation}
    M(\mathbf{u}) = \sum_{k=1}^{K} T_k \alpha_k \text{, where } T_k = \prod_{j=1}^{k-1} (1 - \alpha_j).
\end{equation}
The mask loss is defined as:
\begin{equation}
    L_{msk} = \sum_{\mathbf{u} \in \mathcal{U}'} M(\mathbf{u}) M_{gt}(\mathbf{u}),
\end{equation}
where $\mathbf{U}'$ means the set of pixels of the rendered mask, and $M_{gt}$ is the ground-truth mask of 2D dynamic regions.
This effectively regularizes the optimization of the geometry of dynamic regions by confining it to the visual hulls.

The final loss function is defined as
\begin{equation}
    L = L_{img} + \lambda_{lpips} L_{lpips} + \lambda_{msk} L_{msk},
\end{equation}
where $\lambda_{lpips}$ and $\lambda_{msk}$ are hyperparameters controlling weights of correspondings losses.

\subsection{Inference}
\label{sec:inference}

After training, we apply a few acceleration techniques to boost the rendering speed of our model.
First, we precompute the point location $\mathbf{p}$, radius $r$, density $\sigma$, SH coefficients $\mathbf{s}$ and color blending weights $w_i$ before inference, which are stored at the main memory.
During rendering, these properties are asynchronously streamed onto the graphics card, overlapping rasterization with memory copy to achieve an optimal rendering speed \cite{sanders2010cuda,shreiner2009opengl}.
After applying this technique, the runtime computation is reduced to only a depth peeling evaluation (\cref{sec:rendering}) and a spherical harmonics evaluation (\cref{eq:appearance}).
Second, we convert the model from 32-bit floats to 16-bits for efficient memory access, which increases FPS by 20 and leads to no visible performance loss as validated in \cref{tab:ablation_timing}.
Third, the number of rendering passes $K$ for the differentiable depth peeling algorithm is reduced from 15 to 12, also leading to a 20 FPS speedup with no visual quality change.
Detailed analyses of rendering speed can be found in \cref{sec:ablation}.

\begin{figure*}[t]
    \centering
    \includegraphics[width=1.0\textwidth]{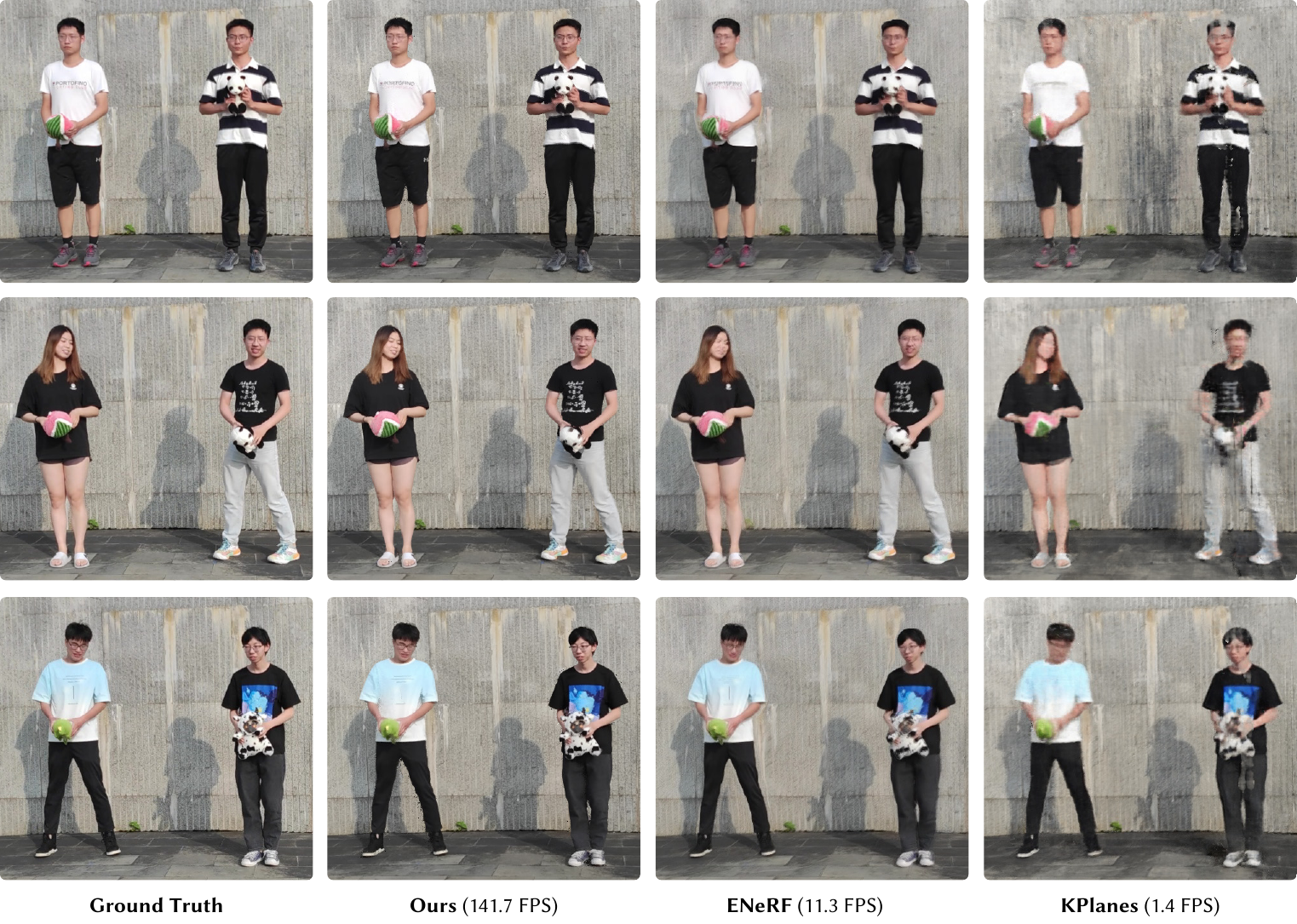}
    \vspace{-17pt}
    \caption{%
        \textbf{Qualitative comparison on the ENeRF-Outdoor \cite{lin2022efficient} dataset that contains 960 $\times$ 540 images.} 
        Our method achieves much higher rendering quality and can be rendered 13$\times$ faster than ENeRF\cite{lin2022efficient}.
        More dynamic results can be found in the supplementary video.
    }
    \label{fig:comparison_enerf_outdoor}
    \vspace{-5pt}
\end{figure*}

\begin{figure*}[t]
    \centering
    \includegraphics[width=1.0\textwidth]{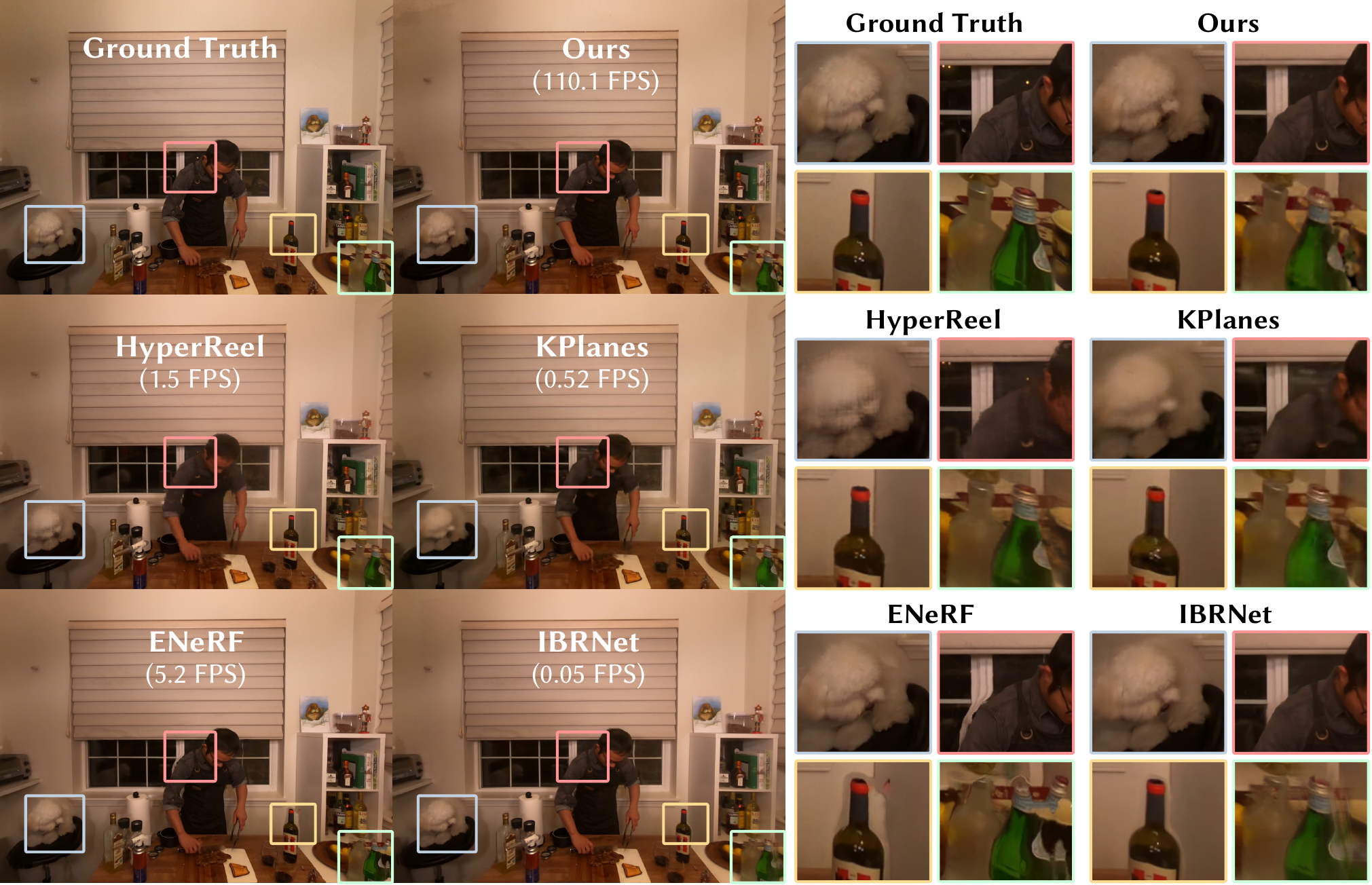}
    \vspace{-17pt}
    \caption{%
        \textbf{Qualitative comparison on the Neural3DV \cite{li2022neural} dataset that contains 1352$\times$1224 images.}
        Our method can not only recover high-frequency details of dynamic objects but also maintain sharp edges around occlusion.
    }
    \label{fig:comparison_neural3dv}
    \vspace{-5pt}
\end{figure*}

\begin{figure*}[t]
    \centering
    \includegraphics[width=1.0\textwidth]{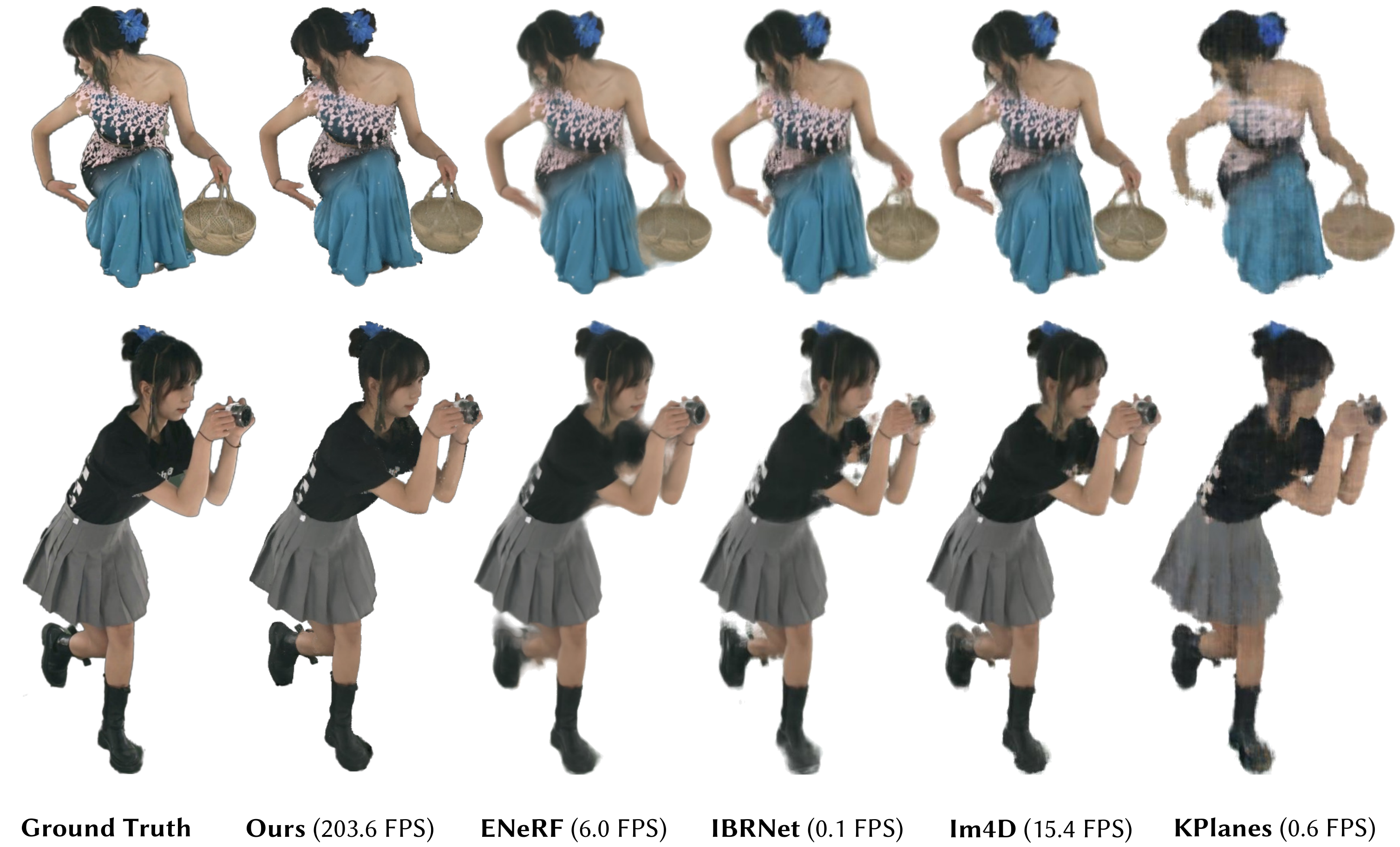}
    \vspace{-17pt}
    \caption{%
        \textbf{Qualitative comparison on the DNA-Rendering \cite{cheng2023dna} dataset that contains 1024$\times$1224 (and 1125$\times$1536) images.}
        Our method can produce high-fidelity images at over 200 FPS while other competitors fail to produce high-quality results for highly dynamic scenes.
    }
    \label{fig:comparison_renbody}
    \vspace{-5pt}
\end{figure*}

\section{Implementation Details}
\label{sec:implementation}

\noindent\textbf{Optimization.} \method is trained using the PyTorch framework \cite{paszke2019pytorch}.
Using the Adam optimizer \cite{kingma2014adam} with a learning rate $5e^{-3}$, our models typically converge after 800k iterations for a sequence length of 200 frames, which takes around 24 hours on a single RTX 4090 GPU.
Specifically, the learning rate of point positions is set to $1e^{-5}$, and the regularization loss weights $\lambda_{lpips}$ and $\lambda_{msk}$ are set to $1e^{-3}$.
During training, the number of passes $K$ for the differentiable depth peeling is set to 15, and the number of nearest input views $N'$ is set to 4.
The rendering speed of our method is reported on an RTX 3090 GPU for the experiments in \cref{sec:experiments} unless otherwise stated.

\vspace{2pt}
\noindent\textbf{Initialization of point clouds.}
We leverage exisiting multi-view reconstruction methods to initialize the point clouds.  %
For dynamic regions, we use segmentation methods \cite{lin2022robust} to obtain their masks in input images and utilize the space carving algorithm \cite{kutulakos2000theory} to extract their coarse geometry.
For static background regions, we leverage foreground masks to compute the mask-weighted average of background pixels along all frames, producing background images without the foreground content.
Then, an Instant-NGP \cite{muller2022instant} model is trained on these images, from which we obtain the initial point clouds.
After the initialization, the number of points for the dynamic regions is typically 250k per frame, and the static background regions typically consist of 300k points.

\section{Experiments}
\label{sec:experiments}

\noindent\textbf{Datasets and metrics.}
We train and evaluate our method \method on multiple widely used multi-view datasets, including DNA-Rendering \cite{cheng2023dna}, ENeRF-Outdoor \cite{lin2022efficient}, NHR \cite{wu2020multi} and Neural3DV \cite{li2022neural}.
DNA-Rendering \cite{cheng2023dna} records 10-second clips of dynamic humans and objects at 15 FPS using 4K and 2K cameras with 60 views.
This dataset is very challenging due to the complex clothing and fast-moving humans recorded.
We conduct experiments on 4 sequences of DNA-Rendering, with 90\% of the views as training set and the rest as evaluation set.
ENeRF-Outdoor \cite{lin2022efficient} records multiple dynamic humans and objects in an outdoor environment at 30FPS using 1080p cameras.
We select three 100-frame sequences with 6 different actors (2 for each sequence) holding objects to evaluate our method \method.
This dataset is difficult for dynamic view synthesis in that not only are there multiple moving humans and objects in one clip, but the background is also dynamic due to the shadow of the humans.
Following Im4D \cite{lin2023im4d} and NeuralBody \cite{peng2021neural}, we evaluate metrics on the dynamic regions for the DNA-Rendering \cite{cheng2023dna} and NHR \cite{wu2020multi} dataset, which can be obtained by predefining the 3D bounding box of the person and projecting it onto the images.
For ENeRF-Outdoor \cite{lin2022efficient}, we jointly train the dynamic geometry and appearance of the foreground and the dynamic appearance of the background to obtain rendering results on the whole image.
All images are resized with a ratio of 0.5 for evaluation and 0.375 if the original resolution is more than 2K.
For DNA-Rendering, the rendered image size is 1024$\times$1224 (and 1125$\times$1536) and for ENeRF-Outdoor, the resolution is 960$\times$540 during the experiments.
The resolutions for Neural3DV video and NHR are 1352$\times$1224 and 512$\times$612 (and 384$\times$512) respectively.
Detailed dataset settings can be found in \cref{sec:dataset_settings}.

\subsection{Comparison Experiments}
\label{sec:comparison}

\begin{table}[t]
    \caption{%
        \textbf{Quantitative comparison on the DNA-Rendering \cite{cheng2023dna} dataset.}
        Image resolutions are 1024$\times$1224 and 1125$\times$1536.
        Metrics are averaged over all scenes.
        Green and yellow cell colors indicate the best and the second best results, respectively.
    }
    \label{tab:comparison_renbody}
    \vspace{-7pt}
    \centering\small
    \setlength{\tabcolsep}{6.5pt}
    \begin{tabular}{lcccc} %
        \toprule
                                      & PSNR $\uparrow$    & SSIM $\uparrow$   & LPIPS $\downarrow$ & FPS                \\
        \midrule
        ENeRF \cite{lin2022efficient} & 28.108             & 0.972             & \cellsecond 0.056  & 6.011              \\
        IBRNet \cite{wang2021ibrnet}  & 27.844             & 0.967             & 0.081              & 0.100              \\
        KPlanes \cite{fridovich2023k} & 27.452             & 0.952             & 0.118              & 0.640              \\
        Im4D~\cite{lin2023im4d}       & \cellsecond 28.991 & \cellsecond 0.973 & 0.062              & \cellsecond 15.360 \\
        \midrule
        Ours                          & \cellfirst 31.173  & \cellfirst 0.976  & \cellfirst 0.055   & \cellfirst 203.610 \\
        \bottomrule
    \end{tabular}
    \vspace{-10pt}
\end{table}

\vspace{2pt}
\noindent\textbf{Comparison results.}
Qualitative and quantitative comparisons on the DNA-Rendering \cite{cheng2023dna} are shown in \cref{fig:comparison_renbody} and \cref{tab:comparison_renbody} respectively.
As evident in \cref{tab:comparison_renbody}, our method \method renders 30x faster than the SOTA real-time dynamic view synthesis method ENeRF \cite{lin2022efficient} with superior rendering quality.
Even when compared with concurrent work \cite{lin2023im4d}, our method \method still achieves 13x speedup and produces consistently higher quality images.
As shown in \cref{fig:comparison_renbody}, KPlanes \cite{fridovich2023k} could not recover the highly detailed appearance and geometry of the 4D dynamic scene.
Other image-based methods \cite{wang2021ibrnet,lin2022efficient,lin2023im4d} produce high-quality appearance.
However, they tend to produce blurry results around occlusions and edges, leading to degradation of the visual quality while maintaining interactive framerate at best.
On the contrary, our method \method can produce higher fidelity renderings at over 200 FPS.
\cref{fig:comparison_enerf_outdoor} and \cref{tab:comparison_enerf_outdoor} provides qualitative and quantitaive results on the ENeRF-Outdoor \cite{lin2022efficient} dataset.
Even on the challenging ENeRF-Outdoor dataset with multiple actors and the background, our method \method still achieves notably better results while rendering at over 140 FPS.
ENeRF \cite{lin2022efficient} produces blurry results on this challenging dataset, and the rendering results of IBRNet \cite{wang2021ibrnet} contain black artifacts around the edges of the images as shown in \cref{fig:comparison_enerf_outdoor}.
K-Planse \cite{fridovich2023k} fails to reconstruct the dynamic humans and varying background regions.
More comparison results on the NHR \cite{wu2020multi} dataset can be found in \cref{sec:additional_experiments}.

\begin{table}[t]
    \caption{%
        \textbf{Quantitative comparison on the ENeRF-Ourdoor \cite{lin2022efficient} dataset.}
        This dataset includes 960 $\times$ 540 images.
        Green and yellow cell colors indicate the best and the second best results, respectively.
    }
    \label{tab:comparison_enerf_outdoor}
    \vspace{-7pt}
    \centering\small
    \setlength{\tabcolsep}{7pt}
    \begin{tabular}{lcccc} %
        \toprule
                                      & PSNR $\uparrow$    & SSIM $\uparrow$   & LPIPS $\downarrow$ & FPS                \\
        \midrule
        ENeRF \cite{lin2022efficient} & \cellsecond 25.452 & 0.809             & 0.273              & \cellsecond 11.309 \\
        IBRNet \cite{wang2021ibrnet}  & 24.966             & \cellfirst 0.929  & \cellsecond 0.172  & 0.140              \\
        KPlanes \cite{fridovich2023k} & 21.310             & 0.735             & 0.454              & 1.370              \\
        \midrule
        Ours                          & \cellfirst 25.815  & \cellsecond 0.898 & \cellfirst 0.147   & \cellfirst 141.665 \\
        \bottomrule
    \end{tabular}
    \vspace{-10pt}
\end{table}

\begin{table}[t]
    \caption{%
        \textbf{Quantitative comparison on the Neural3DV \cite{li2022neural} dataset.}
        This dataset includes 1352$\times$1224 images.
        Green and yellow cell colors indicate the best and the second best results, respectively.
    }
    \label{tab:comparison_neural3dv}
    \vspace{-7pt}
    \centering\small
    \setlength{\tabcolsep}{6.5pt}
    \begin{tabular}{lcccc} %
        \toprule
                                            & PSNR $\uparrow$    & SSIM $\uparrow$   & LPIPS $\downarrow$ & FPS                \\
        \midrule
        ENeRF \cite{lin2022efficient}       & 30.306             & 0.962             & 0.185              & \cellsecond 5.187  \\
        IBRNet \cite{wang2021ibrnet}        & 31.520             & 0.963             & 0.169              & 0.053              \\
        KPlanes \cite{fridovich2023k}       & 31.610             & 0.961             & 0.182              & 0.518              \\
        HyperReel \cite{attal2023hyperreel} & \cellsecond 32.198 & \cellsecond 0.970 & \cellfirst 0.161   & 1.540              \\
        \midrule
        Ours                                & \cellfirst 32.855  & \cellfirst 0.973  & \cellsecond 0.167  & \cellfirst 110.063 \\
        \bottomrule
    \end{tabular}
    \vspace{-10pt}
\end{table}

\subsection{Ablation Studies}
\label{sec:ablation}
\begin{figure*}[t]
    \centering
    \includegraphics[width=1.0\textwidth]{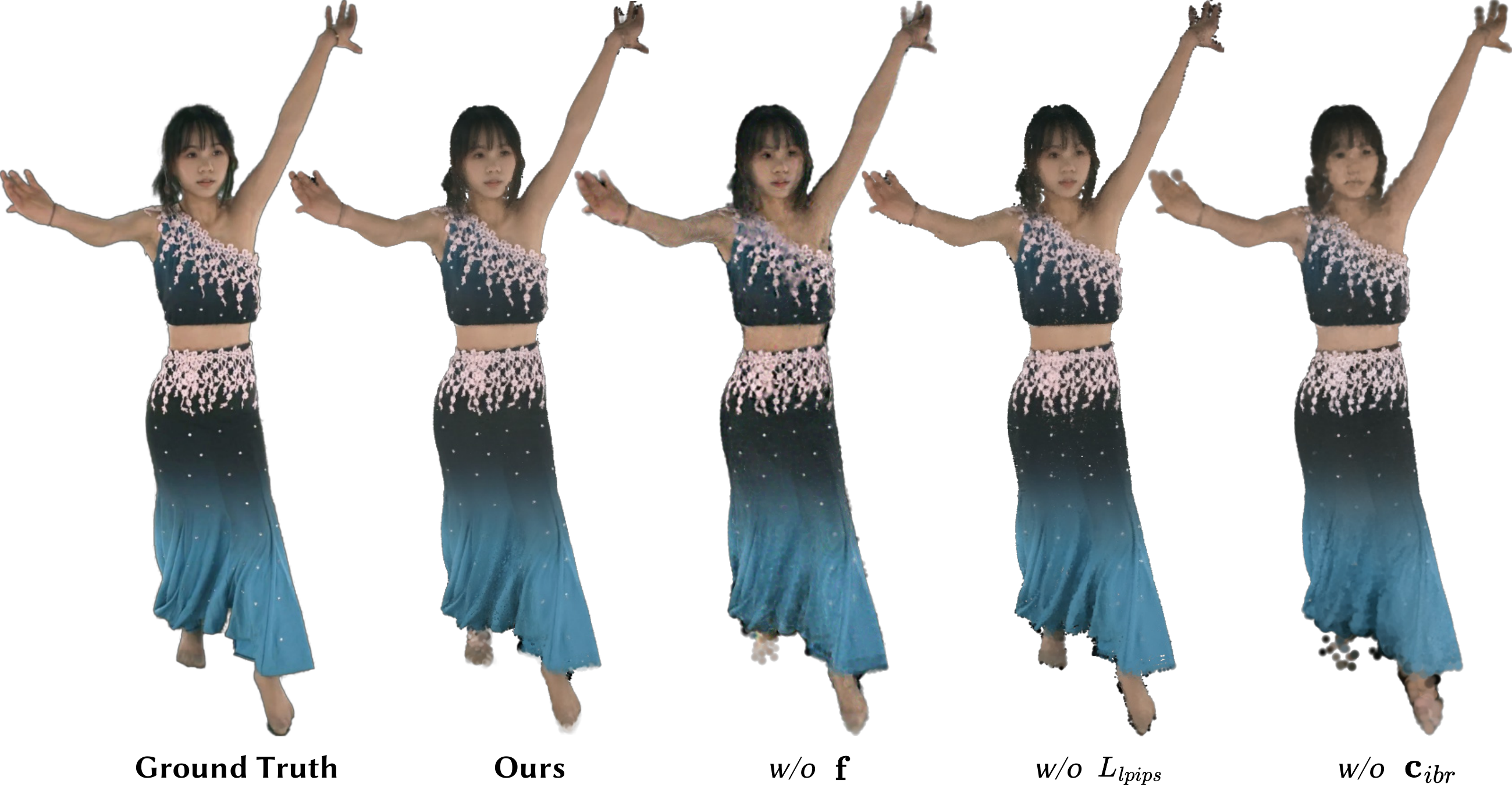}
    \caption{%
        \textbf{Ablation study of proposed components on the \textit{0013\_01} sequence of the DNA-Rendering dataset \cite{cheng2023dna}.}
        Removing our proposed components leads to noisy geometry and blurry appearance.
        Our method produces high-fidelity results with perceptually accurate shapes and colors.
        See \cref{sec:ablation} for more detailed descriptions.
    }
    \label{fig:ablation_module}
\end{figure*}

We perform ablation studies on the 150-frame \textit{0013\_01} sequence of the DNA-Rendering \cite{cheng2023dna} dataset.
Qualitative and quantitative results are shown in \cref{fig:ablation_module} and \cref{tab:ablation_module,tab:ablation_storage,tab:ablation_timing,tab:ablation_resolution}.

\vspace{2pt}
\noindent\textbf{Ablation study on the 4D embedding.}
The ``\textit{w/o} $\mathbf{f}$'' variant removes the proposed 4D embedding (\cref{sec:representation}) module and replaces it with a per-frame and per-point optimizable position, radius, density, and scale.
As shown in \cref{fig:ablation_module} and \cref{tab:ablation_module}, the ``\textit{w/o} $\mathbf{f}$'' variant produces blurry and noisy geometry without the 4D embedding $\Theta$, which leads to the inferior rendering quality.

\vspace{2pt}
\noindent\textbf{Ablation study on the hybrid appearance model.}
The ``\textit{w/o} $\mathbf{c}_{ibr}$'' variant removes $\mathbf{c}_{ibr}$ in the appearance formulation \cref{eq:appearance}, which not only leads to less details on the recovered appearance but also significantly impedes the quality of the geometry.
Adding an additional degree for the SH coefficients does not lead to a significant performance change (PSNR 30.202 \textit{vs.} 30.328).
Comparatively, our proposed method produces high-fidelity rendering with much better details.
A visualization of the view-dependent effect produced by $\mathbf{c}_{sh}$ can be found in \cref{sec:additional_experiments}.

\vspace{2pt}
\noindent\textbf{Ablation study on loss functions.}
As shown in \cref{tab:ablation_module}, removing the $L_{lpips}$ term not only reduces the perceptual quality (LPIPS score) but also leads to the degradation of other performance metrics.
For the highly dynamic DNA-Rendering \cite{cheng2023dna} dataset, the mask loss $L_{msk}$ helps with regularizing the optimization of the dynamic geometry.

\begin{table}[t]
    \caption{%
        \textbf{Ablation studies} on the 150-frame \textit{0013\_01} sequence of the DNA-Rendering dataset \cite{cheng2023dna}.
        ``\textit{w/o} $\mathbf{f}$'' indicates replacing the 4D embedding with a per-frame and per-point optimizable position, radius, density, and scale.
        See \cref{sec:ablation} for more detailed descriptions.
    }
    \label{tab:ablation_module}
    \vspace{-7pt}
    \centering\small
    \setlength{\tabcolsep}{6.5pt}
    \begin{tabular}{lccccc} %
        \toprule
                                        & PSNR $\uparrow$   & SSIM $\uparrow$  & LPIPS $\downarrow$ & Model Size           \\
        \midrule
        \textit{w/o} $\mathbf{f}$       & 29.779            & 0.967            & 0.057              & 1304.0 MiB           \\
        \textit{w/o} $\mathbf{c}_{ibr}$ & 30.259            & 0.973            & 0.054              & \cellfirst 225.0 MiB \\
        \textit{w/o} $\mathbf{c}_{sh}$  & 31.946            & 0.981            & \cellfirst 0.040   & \cellfirst 225.0 MiB \\
        \textit{w/o} $L_{lpips}$        & 31.661            & 0.979            & 0.063              & \cellfirst 225.0 MiB \\
        \textit{w/o} $L_{msk}$          & 29.115            & 0.965            & 0.073              & \cellfirst 225.0 MiB \\
        \midrule
        Ours                            & \cellfirst 31.990 & \cellfirst 0.982 & \cellfirst 0.040   & \cellfirst 225.0 MiB \\
        \bottomrule
    \end{tabular}
    \vspace{-10pt}
\end{table}
\begin{table*}[t]
    \caption{%
        \textbf{Storage analysis of our method on the 150-frame \textit{0013\_01} sequence of the DNA-Rendering dataset \cite{cheng2023dna}.}
        ``Storage'' indicates the size of model stored on disk, and ``Storage / Frame'' indicates the per-frame size.
    }
    \label{tab:ablation_storage}
    \centering\footnotesize
    \setlength{\tabcolsep}{6.3pt}
    \begin{tabular}{lcccc|cc} %
        \toprule
                        & Point Positions $\mathbf{p}$ & 4D Embedding $\Theta$ & MLPs and CNNs & Total Model Size & Encoded Video & Total Storage (\textit{w/} Videos) \\
        \midrule
        Storage         & 208.09 MB                   & 16.77 MB             & 0.10 MB      & 224.96 MB       & 62.89 MB     & 287.86 MB                         \\
        Storage / Frame & 1.387  MB                   & 0.112  MB            & 0.001 MB     & 1.500 MB        & 0.419 MB     & 1.919 MB                          \\
        \bottomrule
    \end{tabular}
\end{table*}

\vspace{2pt}
\noindent\textbf{Storage analysis.}
For the 150-frame \textit{0013\_01} scene, the storage analysis of our method \method is listed in \cref{tab:ablation_storage}.
The point positions $\mathbf{p}$ take up the majority of the model size due to its explicit representation.
The final storage cost for our method is less than 2 MB per frame with source videos included.
The input images of DNA-Rendering \cite{cheng2023dna} are provided in JPEG formats.
We encode frames of all input images as videos using the HEVC encoder of FFmpeg with a CRF of 25 \cite{tomar2006converting}.
After the encoding, we obverse no change in LPIPS (0.040), no loss in SSIM (0.982), and only a 0.42\% decrease in PSNR (31.990 \textit{vs.} 31.855), which indicates that our method \method is robust to the video encoding of input images.
After encoding the input images as videos, the storage overhead for Image-Based Rendering (\cref{sec:representation}) is only 0.419 MB per frame with minimal rendering quality change.

As mentioned in \cref{sec:inference}, we precompute the physical properties on the point clouds for real-time rendering, which takes around 2 seconds for one frame.
Although large in size (200 MB for one frame of \textit{0013\_01}), these precomputed caches only reside in the main memory and are not explicitly stored on disk, which is feasible for a modern PC.
This makes our representation a form of compression, where the disk file size is small (2 MB per frame) but the information contained is very rich (200 MB per frame).

\subsection{Rendering Speed Analysis}
\begin{table}[t]
    \caption{%
        \textbf{Runtime analysis of the proposed method on the \textit{0013\_01} sequence of DNA-Rendering \cite{cheng2023dna}.}
        The acceleration techniques all lead to minimal quality changes as shown by the cell coloring (green for the best and yellow for the second best).
        See \cref{sec:ablation} for detailed descriptions.
    }
    \label{tab:ablation_timing}
    \vspace{-7pt}
    \centering\small
    \setlength{\tabcolsep}{7pt}
    \begin{tabular}{lcccc} %
        \toprule
                              & PSNR $\uparrow$    & SSIM $\uparrow$  & LPIPS $\downarrow$ & FPS                 \\
        \midrule
        \textit{w/o} fp16     & \cellfirst 32.020  & \cellfirst 0.982 & \cellfirst 0.039   & \cellsecond 202.021 \\
        \textit{w/o} $K = 12$ & 31.951             & \cellfirst 0.982 & \cellsecond 0.040  & 200.397             \\
        \textit{w/o} Cache    & 31.969             & \cellfirst 0.982 & \cellsecond 0.040  & 22.193              \\
        \textit{w/o} DDP      & 31.900             & 0.981            & 0.041              & 29.656              \\
        \midrule
        Ours                  & \cellsecond 31.990 & \cellfirst 0.982 & \cellsecond 0.040  & \cellfirst 219.430  \\
        \bottomrule
    \end{tabular}
    \vspace{-10pt}
\end{table}
\begin{table}[t]
    \caption{%
        \textbf{Rendering speed on different GPUs and resolutions.}
        The results are recorded on the first frame of the \textit{0013\_01} sequence of DNA-Rendering \cite{cheng2023dna} and the \textit{actor1\_4} sequence of ENeRF-Outdoor \cite{lin2022efficient} with the interactive GUI.
        Resolutions are set to 720p (720 $\times$ 1280), 1080p (1080 $\times$ 1920), and 4K (2160 $\times$ 3840).
        Even with the overhead of the interactive GUI (``\textbf{\textit{w/} GUI}''), our method still achieves unprecedented rendering speed.
        More real-time rendering results can be found in the supplementary video.
    }
    \label{tab:ablation_resolution}
    \vspace{-7pt}
    \centering\small
    \setlength{\tabcolsep}{2pt}
    \begin{tabularx}{0.475\textwidth}{Xlccc} %
        \toprule
        Dataset                                                & Res.  & RTX 3060  & RTX 3090  & RTX 4090  \\
        \midrule
        \multirow{3}{*}{\shortstack{DNA-Rendering \cite{cheng2023dna} \\ \textbf{\textit{w/} GUI}}}     & 720p  & 173.8 FPS & 246.9 FPS & 431.0 FPS \\
                                                               & 1080p & 138.7 FPS & 233.1 FPS & 409.8 FPS \\
                                                               & \textbf{4K}    & \textbf{90.0 FPS}  & \textbf{147.4 FPS} & \textbf{288.8 FPS} \\
        \midrule
        \multirow{3}{*}{\shortstack{ENeRF-Outdoor \cite{lin2022efficient} \\ \textbf{\textit{w/} GUI}}} & 720p  & 90.5 FPS  & 130.5 FPS & 351.5 FPS \\
                                                               & 1080p & 66.1 FPS  & 103.6 FPS  & 249.7 FPS \\
                                                               & \textbf{4K}    & \textbf{25.1 FPS}  & \textbf{47.2 FPS}  &\textbf{85.1 FPS}  \\
        \bottomrule
    \end{tabularx}
    \vspace{-20pt}
\end{table}

As mentioned in \cref{sec:inference}, we introduce a number of optimization techniques to accelerate the rendering speed of our method \method, which are only made possible by our proposed hybrid geometry and appearance representation.
In \cref{tab:ablation_timing}, we ablate the effectiveness and quality impact of those proposed techniques on the 150-frame \textit{0013\_01} sequence of the DNA-Rendering \cite{cheng2023dna} dataset.

\vspace{2pt}
\noindent\textbf{The effectiveness of precomputation.}
For real-time rendering, we precompute and cache $\mathbf{p}$, $r$, $\sigma$ and $\mathbf{s}$ for all points and store them in the main memory.
Thanks to our design choice of splitting the appearance representation into constant $\mathbf{c}_{ibr}$ and view-dependent $\mathbf{c}_{sh}$, we can also precompute and cache the per-image weights $w$ and color $\mathbf{c}_{img}$ for all source images (\cref{sec:representation}).
These caches take around 200MB per frame of main memory for the 150-frame 60-view scene of \textit{0013\_01} of the DNA-Rendering \cite{cheng2023dna} dataset.
The pre-computation enabled by our representation (\cref{sec:representation}) achieves a 10x speedup (Ours \textit{vs.} ``\textit{w/o} Cache'').

\vspace{2pt}
\noindent\textbf{Differentiable depth peeling.}
We also make comparisons with more traditional CUDA-based differentiable point cloud rendering technique provided by PyTorch3D \cite{ravi2020pytorch3d} (``\textit{w/o} DDP'') to validate the effectiveness of our proposed differentiable depth peeling algorithm (\cref{sec:rendering}).
Both our proposed DDP (\cref{sec:rendering}) and PyTorch3D's \cite{ravi2020pytorch3d} implementation use the same volume rendering equation as in \cref{eq:volume_rendering}.
As shown in \cref{tab:ablation_timing}, our proposed method is more than 7 times faster than the CUDA-based one.

\vspace{2pt}
\noindent\textbf{Other acceleration techniques.}
The ``\textit{w/o} fp16'' variant uses the original 32-bit floating point number for computation.
The ``\textit{w/o} $K=12$'' variant uses 15 passes in the depth peeling algorithm as when training.
Using 16-bit floats and 12 rendering passes both lead to a 20FPS speedup.

\vspace{2pt}
\noindent\textbf{Rendering speed on different GPUs and resolutions.}
We additionally report the rendering speed of our method on different hardware (RTX 3060, 3090, and 4090) with different resolutions (720p, 1080p, and 4K (2160p)) in \cref{tab:ablation_resolution}.
The rendering speed reported here contains the overhead of the interactive GUI (``\textbf{\textit{w/} GUI}''), making them slightly slower than those reported in \cref{sec:comparison}.
\method achieves real-time rendering speed even when rendering 4K (2160p) images on commodity hardware as shown in the table.

\section{Conclusion and Discussion}
In this paper, we provide a neural point cloud-based representation, \method, for real-time rendering of dynamic 3D scenes at 4K resolution.
We build \method upon a 4D feature grid to naturally regularize the points and develop a novel hybrid appearance model for high-quality rendering.
Furthermore, we develop a differentiable depth peeling algorithm that utilizes the hardware rasterization pipeline to effectively optimize and efficiently render the proposed model.
In our experiments, we show that \method not only achieves state-of-the-art rendering quality but also exhibits a more than 30$\times$ increase in rendering speed (over 200FPS at 1080p on an RTX 3090).

However, our method still has some limitations.
\method cannot produce correspondences of points across frames, which are important for some downstream tasks.
Moreover, the storage cost for \method increases linearly with the number of video frames, so our method has difficulty in modeling long volumetric videos.
How to model correspondences and reduce the storage cost for long videos could be two interesting problems for future works.

{\small
    \bibliographystyle{ieee_fullname}
    \bibliography{reference}
}

\ifarxiv
    \clearpage
    \appendix
\newcommand{\AppendixPrefix}{A}
\renewcommand{\thefigure}{\AppendixPrefix\arabic{figure}}
\setcounter{figure}{0}
\renewcommand{\thetable}{\AppendixPrefix\arabic{table}}
\setcounter{table}{0}
\renewcommand{\theequation}{\AppendixPrefix\arabic{equation}}
\setcounter{equation}{0}
\section*{Appendix}

\section{Dataset Settings}
\label{sec:dataset_settings}

\begin{table}[t]
    \caption{%
        \textbf{Number of views and frames for each dataset's used sequences.}
        The \textit{basketball} sequence of NHR \cite{wu2020multi} provides 72 views compared to the 56 views for the rest of the dataset.
    }
    \label{tab:dataset_setting}
    \vspace{-7pt}
    \centering\small
    \setlength{\tabcolsep}{2pt}
    \begin{tabularx}{0.475\textwidth}{l *4{>{\centering\arraybackslash}X}@{}} %
        \toprule
        Dataset                                      & Sequence\newline Count & Traininig\newline View & Testing\newline View & Frame\newline Count \\
        \midrule
        DNA-Rendering \cite{cheng2023dna}            & 4                      & 56                     & 4                    & 150                 \\
        NHR (\textit{sport}) \cite{wu2020multi}      & 3                      & 52                     & 4                    & 100                 \\
        NHR (\textit{basketball}) \cite{wu2020multi} & 1                      & 68                     & 4                    & 100                 \\
        ENeRF-Outdoor \cite{lin2022efficient}        & 3                      & 17                     & 1                    & 100                 \\
        Neural3DV \cite{li2022neural}                & 1                      & 19                     & 1                    & 300                 \\
        \bottomrule
    \end{tabularx}
    \vspace{-10pt}
\end{table}

In \cref{tab:dataset_setting}, we list the detailed sequence count, view count, and frame count for each of the datasets used in our experiments.
For image-based baselines \cite{wang2021ibrnet,lin2022efficient} and our method, the training view is also the source view for the IBR process.
No test views are provided as source views.

\section{Additional Experiments}
\label{sec:additional_experiments}

\subsection{Additional Comparisons on NHR\cite{wu2020multi}}
\begin{figure*}[t]
    \centering
    \includegraphics[width=1.0\textwidth]{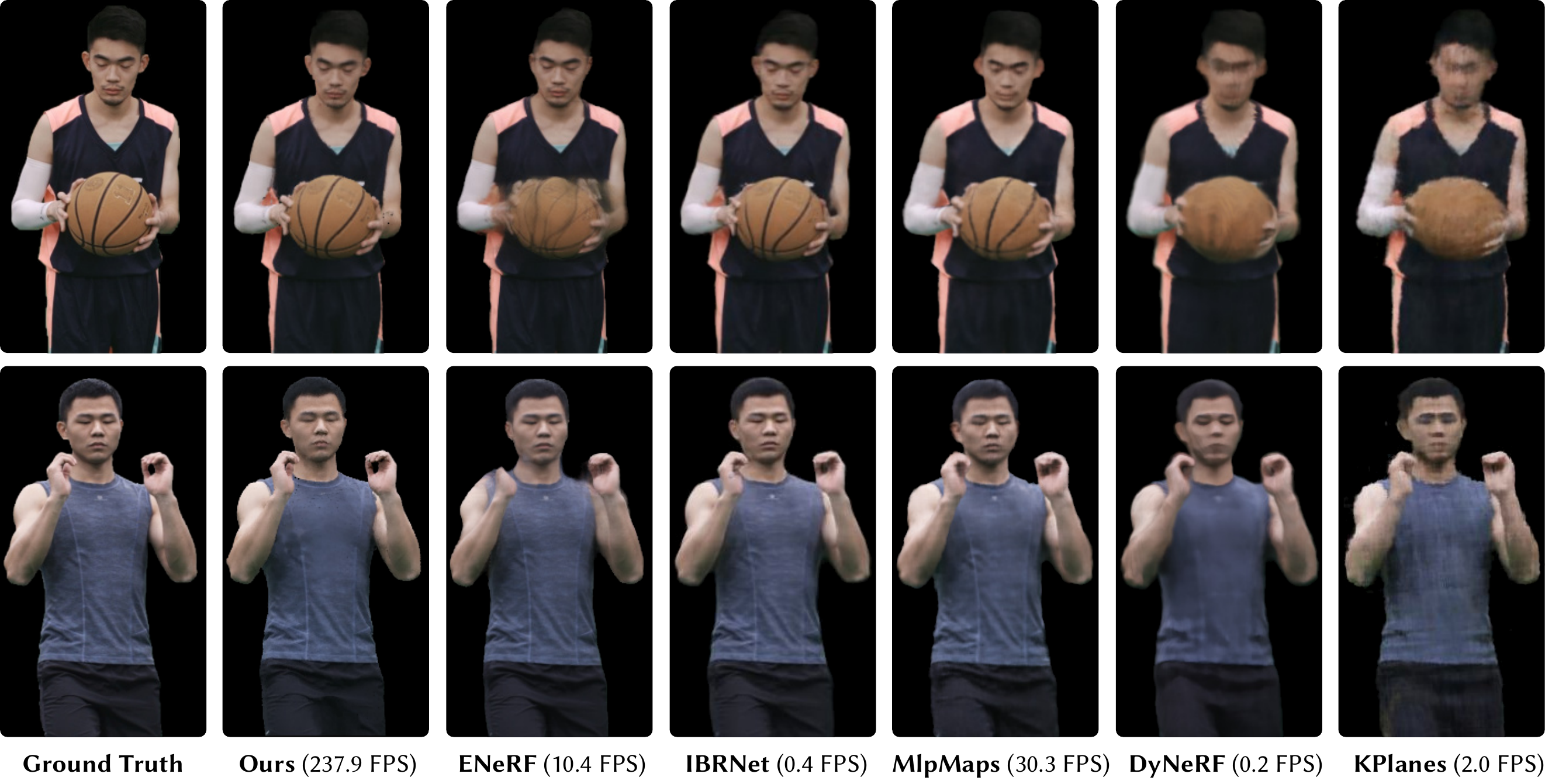}
    \caption{%
        \textbf{Qualitative comparison on the NHR \cite{wu2020multi} dataset that contains 512$\times$612 (and 384$\times$512) images.} 
    }
    \label{fig:comparison_nhr}
    \vspace{-5pt}
\end{figure*}

\begin{table}[t]
    \caption{%
        \textbf{Quantitative comparison on the NHR \cite{wu2020multi} dataset.}
        This dataset includes 512$\times$612 and 384$\times$512 images.
        Metrics are averaged over all scenes.
    }
    \label{tab:comparison_nhr}
    \vspace{-7pt}
    \centering\small
    \setlength{\tabcolsep}{6.5pt}
    \begin{tabular}{lcccc} %
        \toprule
                                            & PSNR $\uparrow$    & SSIM $\uparrow$   & LPIPS $\downarrow$ & FPS                \\
        \midrule
        ENeRF \cite{lin2022efficient}       & 30.765             & 0.954             & \cellsecond 0.052  & 10.432             \\
        IBRNet \cite{wang2021ibrnet}        & \cellsecond 33.537 & \cellsecond 0.965 & 0.078              & 0.369              \\
        KPlanes \cite{fridovich2023k}       & 32.933             & 0.958             & 0.101              & 1.979              \\
        MlpMaps \cite{peng2023representing} & 32.203             & 0.953             & 0.080              & \cellsecond 30.303 \\
        DyNeRF \cite{li2022neural}          & 30.872             & 0.943             & 0.117              & 0.192              \\
        \midrule
        Ours                                & \cellfirst 33.743  & \cellfirst 0.973  & \cellfirst  0.045  & \cellfirst 237.919 \\
        \bottomrule
    \end{tabular}
    \vspace{-10pt}
\end{table}
In \cref{tab:comparison_nhr,fig:comparison_nhr}, we provide quantitative and qualitative comparisons on the NHR dataset \cite{wu2020multi} with multiple baselines.
KPlanes \cite{fridovich2023k} and DyNeRF \cite{li2022neural} can only produce blurry images on this dataset that include fast motions, and their rendering speed is also limited (2.0 and 0.2 FPS).
ENeRF \cite{lin2022efficient} and MlpMaps \cite{peng2023representing} can be rendered at interactive frame rates with moderate resolution, but our method can produce higher quality results at much higher frame rates (30FPS \textit{vs.} 238FPS).

\subsection{Visualization of SH color}
\begin{figure*}[t]
    \centering
    \includegraphics[width=1.0\textwidth]{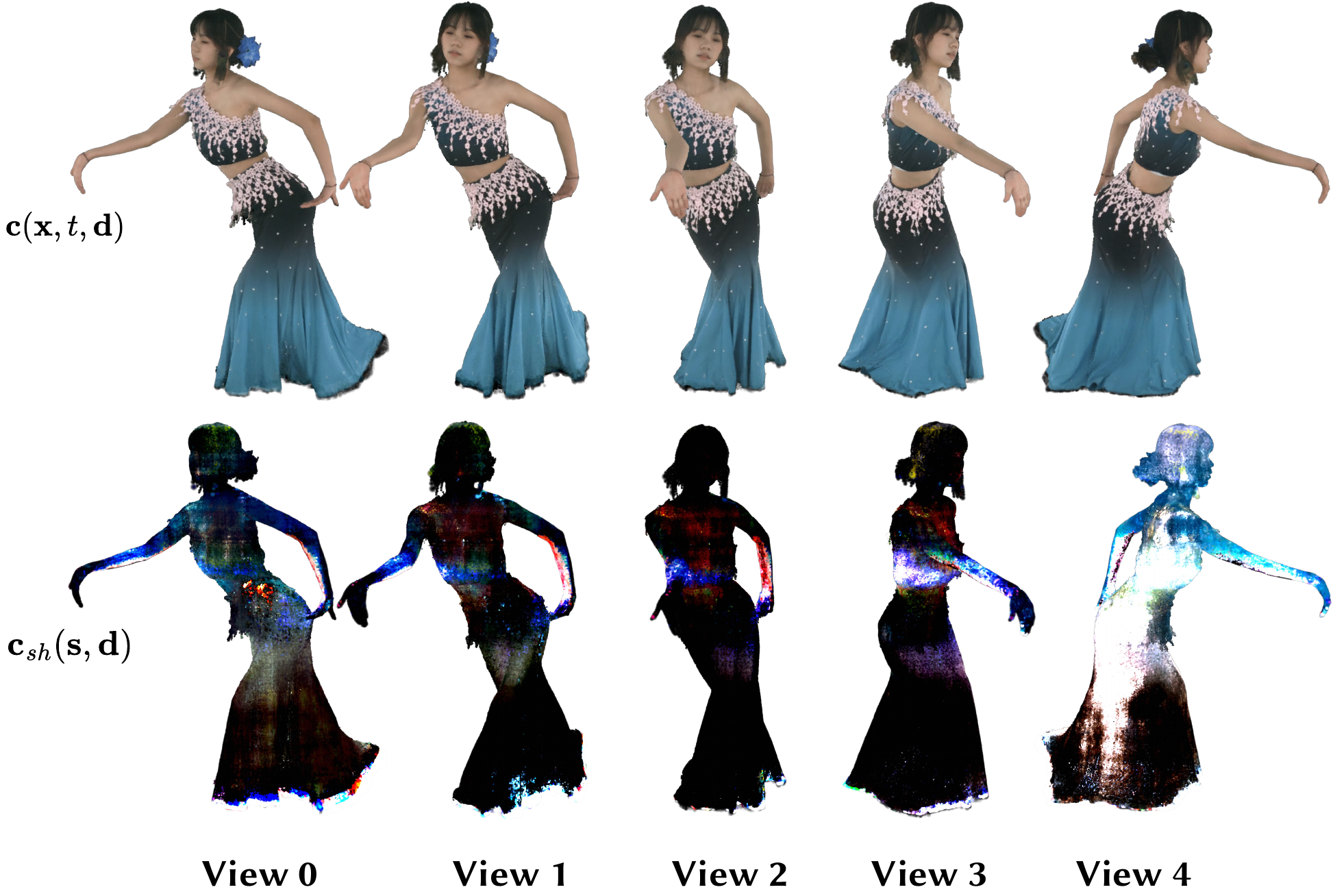}
    \caption{%
        \textbf{Visualization of $\mathbf{c}(\mathbf{x},t,\mathbf{d})$ and $\mathbf{c}_{sh}(\mathbf{s},\mathbf{d})$ on 5 rotating views of the \textit{0013\_01} sequence of DNA-Rendering \cite{cheng2023dna}.}
        The view-dependent SH color $\mathbf{c}_{sh}$ compensates the high-quality but discrete IBR color $\mathbf{c}_{ibr}$.
        We increase the brightness of $\mathbf{c}_{sh}$ for a clearer visualization.
        Details of the implementation can be found in \cref{sec:representation}.
    }
    \label{fig:visualization_sh}
    \vspace{-5pt}
\end{figure*}

In \cref{fig:visualization_sh}, we visualize the SH color $\mathbf{c}_{sh}(\mathbf{s},\mathbf{d})$ on 5 rotating views of the \textit{0013\_01} sequence of DNA-Rendering \cite{cheng2023dna}.
The SH model provides the fine-level appearance that enables the continuous view-dependent effects.

\subsection{Comparisons with 3DGS \cite{kerbl20233d}}

\begin{table}[t]
    \caption{%
        \textbf{Quantitative comparison on the first frame of the all three sequences of the ENeRF-Outdoor \cite{lin2022efficient} dataset.}
        The dataset contains 18 images of 960 $\times$ 540 resolution.
        ``Storage'' indicates the disk file size of the trained models (and source images for our method).
    }
    \label{tab:comparison_static}
    \vspace{-7pt}
    \centering\footnotesize
    \setlength{\tabcolsep}{2pt}
    \begin{tabular}{lcccccc} 
        \toprule
                                    & PSNR $\uparrow$   & SSIM $\uparrow$  & LPIPS $\downarrow$ & FPS                & Storage            & Training            \\
        \midrule
        Gaussian \cite{kerbl20233d} & 21.633            & 0.608            & 0.349              & 88.355             & 715 MB             & \cellfirst 0.5 hour \\
        \midrule
        Ours                        & \cellfirst 26.544 & \cellfirst 0.907 & \cellfirst 0.145   & \cellfirst 148.581 & \cellfirst 16.0 MB & 1.5 hour            \\
        \bottomrule
    \end{tabular}
    \vspace{-10pt}
\end{table}
\begin{figure*}[t]
    \centering
    \includegraphics[width=1.0\textwidth]{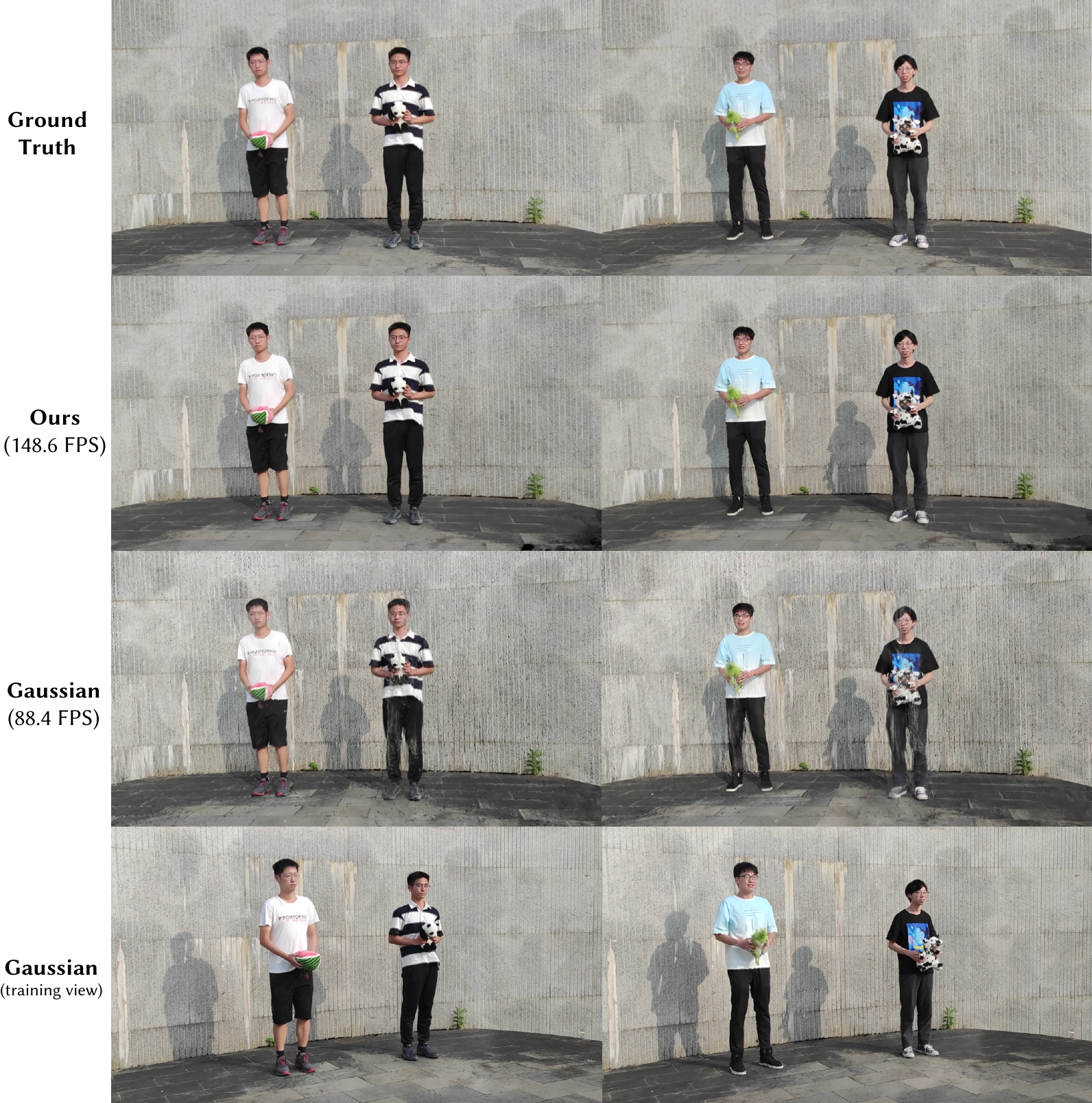}
    \caption{%
        \textbf{Qualitative comparison on the first frame of the \textit{actor1\_4} sequence of ENeRF-Outdoor \cite{lin2022efficient} dataset.}
        The first frame contains 18 images of 960 $\times$ 540 resolution.
        3D Gaussian Splatting \cite{kerbl20233d} overfits the training view as indicated by the last two rows.
    }
    \label{fig:comparison_static}
    \vspace{-5pt}
\end{figure*}

We perform additional comparisons with 3DGS \cite{kerbl20233d} on the first frame of the \textit{actor1\_4} sequence of the ENeRF-Outdoor \cite{lin2022efficient} dataset in \cref{tab:comparison_static,fig:comparison_static}.
The storage cost of our method contains both the file sizes (in MB) of the trained model and source images.
After the precomputation in \cref{sec:inference}, the main memory usage of our method is 100 MB per frame, which is still smaller than the disk file size of 3DGS.
Our method is trained for 5,000 iterations, and 3DGS is trained for 30,000 iterations (their default settings) using their official source code.
3DGS renders slower than our method because too many points were generated during their training process.
Qualitatively, 3DGS overfits the first frame and fails to generalize to novel views, as indicated by the last two rows of \cref{fig:comparison_static}.

Compared to training a 3DGS for every frame, our method is superior in the following ways.
First, the storage cost for 3DGS is too large for even a 100-frame video (715 MB per frame), while our method maintains a reasonable 2 MB per frame storage overhead (\cref{sec:ablation}).
Thanks to the implicit compression of our 4D feature grid and IBR model \cref{sec:ablation}, our method better utilizes the temporal redundancy of the dynamic 3D scene.
The optimization and precomputation of our method can be viewed as a form of compression and decompression, where we encode and decode the dynamic 3D scene using 4D feature grids, network weights, and source videos.
Second, the hybrid image-based appearance model (\cref{sec:representation}) of our method is more expressive than the spherical harmonics utilized by 3DGS, thus achieving higher rendering quality as shown by \cref{fig:comparison_static,tab:comparison_static}.

\fi

\end{document}